\documentclass[runningheads]{llncs}

\usepackage[mobile]{eccv}

\usepackage{eccvabbrv}

\usepackage{graphicx}
\usepackage{booktabs}
\usepackage{amsmath}
\usepackage{amssymb}
\usepackage{float}
\usepackage{algorithm}
\usepackage{algpseudocode}
\usepackage{multirow} 
\usepackage{wrapfig}

\usepackage{pifont}

\usepackage{colortbl}
\usepackage{xcolor}

\usepackage{pifont}
\newcommand{\cmark}{\ding{51}}%
\newcommand{\xmark}{\ding{55}}%

\renewcommand{\paragraph}[1]{\vspace{1.25mm}\noindent\textbf{#1}}

\definecolor{baselinecolor}{gray}{.95}

\makeatletter
\def\blfootnote{\xdef\@thefnmark{}\@footnotetext}
\makeatother

\usepackage[accsupp]{axessibility}  

\usepackage{hyperref}

\usepackage{orcidlink}

\begin{document}
\title{Leveraging Near-Field Lighting for Monocular Depth Estimation from Endoscopy Videos} 
 \titlerunning{PPSNet}
 
\author{Akshay Paruchuri$^{\dagger}$ \and
Samuel Ehrenstein \and
Shuxian Wang \and
Inbar Fried \and
Stephen M. Pizer \and
Marc Niethammer \and
Roni Sengupta$^{\dagger}$}

\authorrunning{Paruchuri et al.}

\institute{Department of Computer Science\\University of North Carolina at Chapel Hill}

\maketitle

\blfootnote{$^{\dagger}$Corresponding authors: \{akshay, ronisen\}@cs.unc.edu}

\begin{abstract}

Monocular depth estimation in endoscopy videos can enable assistive and robotic surgery to obtain better coverage of the organ and detection of various health issues. Despite promising progress on mainstream, natural image depth estimation, techniques perform poorly on endoscopy images due to a lack of strong geometric features and challenging illumination effects. In this paper, we utilize the photometric cues, i.e., the light emitted from an endoscope and reflected by the surface, to improve monocular depth estimation. We first create two novel loss functions with supervised and self-supervised variants that utilize a per-pixel shading representation. We then propose a novel depth refinement network (PPSNet) that leverages the same per-pixel shading representation. Finally, we introduce teacher-student transfer learning to produce better depth maps from both synthetic data with supervision and clinical data with self-supervision. We achieve state-of-the-art results on the C3VD dataset while estimating high-quality depth maps from clinical data. Our code, pre-trained models, and supplementary materials can be found on our project page: \url{https://ppsnet.github.io/}.
  \keywords{Monocular depth estimation \and Photometric refinement \and Sim2Real transfer learning \and Endoscopic imaging}
\end{abstract}

\vspace{-0.5em}
\section{Introduction}
\vspace{-0.5em}

Endoscopy~\cite{berci2000history} is a minimally invasive imaging technique where a slender and flexible tube equipped with a light and a camera is used to inspect internal organs and tissues of the body via an opening, e.g., the mouth or anus. Endoscopy is routinely used for cancer screening, e.g., in colons~\cite{kuipers2013colorectal} and lungs~\cite{andolfi2016role}; for identifying potential issues in the digestive tract, e.g., ulcers, polyps, and hemorrhoids~\cite{kaminski2017performance}; and for respiratory problems, e.g., infections, abnormal narrowing of the airway, and collapsed lungs~\cite{faro2015official}. In 2022 an estimated 225.3 million endoscopic procedures were performed globally with an annual growth rate of 1.3\%~\cite{GrandViewResearch2023}.  Endoscopy is challenging as it requires highly skilled personnel navigating through narrow and winding pathways in the body. As a result, patients may experience discomfort and some regions may remain unsurveyed. Thus researchers have focused on developing intelligent techniques for better guidance and visualization of unsurveyed regions during the procedure, to obtain better coverage of the organ, and to facilitate autonomous navigation. However, these intelligent techniques are still clinically infeasible as they all share the fundamental problem of accurately estimating the 3D organ geometry. 3D understanding of organ shapes will also enable automatic measurement of geometric properties, e.g., measuring the airway cross-sectional area in children to detect any abnormalities without exposing them to radiation~\cite{heimann2009statistical}.

Current attempts~\cite{kaufman20083d, zhao2016endoscopogram, mahmoud2017orbslam, Scarzanella2017DeepM3, widya20193d, bae2020deep, ma2021rnnslam, rau2023simcol3d, zhang2021colde} at 3D understanding of internal organs from endoscopy videos heavily rely on learning priors from synthetic training data and leveraging geometric cues from camera motion. While this leads to a reasonable 3D understanding on synthetic data~\cite{rau2023simcol3d, Bobrow_2023} and imagery captured with a cellphone camera in typical outdoor and indoor environments~\cite{ke2023repurposing, yang2024depth}, it often fails on real clinical data due to the lack of strong geometric features, strong textural cues, strong reflections from mucus layers, and complex illumination effects. This issue is even more pronounced for oblique and en-face views where the camera is looking away from the tubular structures of the colon or the airway, which is often crucial for detecting cancer and other health screenings.

Our key idea is to model the reflection of co-located light with the camera of the endoscope from the surface of the internal organ to help predict the relative distance between the endoscope and the surface. We rely on the fact that any point on the surface closer to the camera and facing the endoscope will reflect more light than a point that is further away or facing away from the endoscope. Thus we introduce a monocular depth estimation algorithm that can leverage near-field illumination and learn from synthetic data with ground-truth supervision and real clinical data with self-supervision. While our method is general for any endoscopy videos, in this work we focus on colonoscopy. This is because colonoscopy screening~\cite{nierengarten2023colonoscopy} is the gold standard for colorectal cancer detection, the third most common cancer in the world, and because datasets of both synthetic (phantom~\cite{Bobrow_2023} and computer graphics generated~\cite{rau2023simcol3d}) and real colonoscopies~\cite{ma2021colon10k, Azagra_2023, wang2024structurepreservingimagetranslationdepth} are publicly available.

\begin{figure}[tb]
  \centering
  \includegraphics[width=\linewidth]{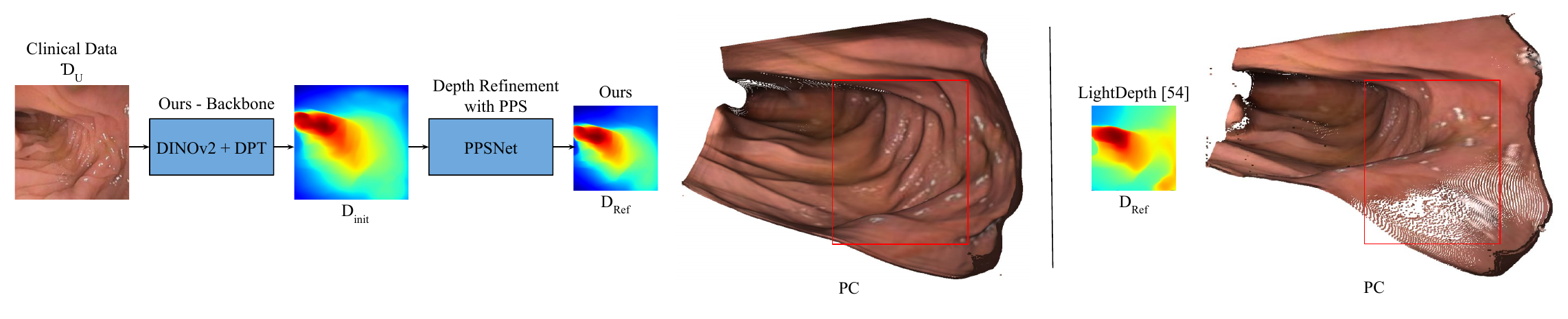}
  \vspace{-2em}
  \caption{Our approach models near-field lighting, emitted by the endoscope and reflected by the surface, as Per-Pixel Shading (PPS). We use PPS feature to perform depth refinement (PPSNet) on clinical data using teacher-student transfer learning and a PPS-informed self-supervision. Our method outperforms the state-of-the-art monocular depth estimation technique LightDepth~\cite{rodríguezpuigvert2023lightdepth}, which trains using self-supervision that explicitly reconstructs input images using illumination decline.}
  \label{fig:main-teaser}
  \vspace{-2em}
\end{figure}

We propose multiple key contributions that harness the power of near-field illumination in predicting depth information. $\bullet$ We first utilize the per-pixel lighting representation, introduced in~\cite{lichy2022fast}, to compute the Per-Pixel Shading (PPS) map which is strongly correlated with the image intensity field. In contrast to ~\cite{lichy2022fast} which only uses per-pixel lighting representation as an input to a surface normal estimation network, we calculate a PPS map and propose a supervised and a self-supervised loss function for training on synthetic and real data respectively. $\bullet$ We then propose a depth refinement architecture, PPSNet, that modulates depth features computed by a monocular depth estimation foundation model, combining DINOv2~\cite{oquab2024dinov2} features and an initial depth from a DPT-based~\cite{Ranftl_2021_ICCV} decoder, fine-tuned on the C3VD~\cite{Bobrow_2023} dataset, with features computed from the PPS map. To our knowledge, existing depth refinement networks~\cite{kusupati2020normal, bae2022irondepth, lichy2021shape} often rely on geometric consistency or uncertainty estimation, whereas we exploit illumination features which are more useful in endoscopy videos. $\bullet$ Finally, we propose a teacher-student transfer learning approach, where the teacher network, effectively a monocular depth estimation foundation model followed by PPSNet refinement, is trained on both synthetic data with supervision and real clinical data with weak supervision. The Teacher additionally guides the learning of the Student network on unlabeled real clinical data with our proposed self-supervised loss function that utilizes PPS representation. In contrast to teacher-student learning-based monocular depth estimation algorithm ~\cite{yang2024depth} that trains on $\sim$60 million unlabeled images with semantic self-supervision, our method is trained with only $\sim$18,000 clinical endoscopy images using lighting based self-supervision.

We show that our proposed approach significantly outperforms state-of-the-art monocular depth estimation techniques on the synthetic C3VD dataset~\cite{Bobrow_2023} both quantitatively and qualitatively. Similar observations also hold true for real clinical datasets~\cite{ma2021colon10k,wang2024structurepreservingimagetranslationdepth}, which we show qualitatively in the absence of any ground-truth depth maps. We perform a detailed ablation study to show the effectiveness of our per-pixel shading-based loss functions, depth refinement modules, and our teacher-student-based transfer learning approach.

In summary, the key contributions of our approach are: $\bullet$ We introduce a supervised and a self-supervised loss function that utilizes near-field illumination emitted by the endoscope and reflected by the surface. $\bullet$ We propose a depth-refinement architecture, PPSNet, that utilizes near-field illumination. $\bullet$ We develop a teacher-student transfer learning approach where a teacher model guides the learning of a student model using our proposed self-supervised loss function. $\bullet$ We perform an extensive evaluation on synthetic C3VD~\cite{Bobrow_2023} and real clinical data~\cite{ma2021colon10k,wang2024structurepreservingimagetranslationdepth}, producing state-of-the-art results. We will release all codes, data, and benchmarks for future research.

\vspace{-0.5em}
\section{Related Works}
\vspace{-0.5em}

\noindent \textbf{Monocular Depth Estimation.} Monocular depth estimation has received a great deal of attention due to its importance in 3D understanding from a single image. Researchers have considered various different approaches ranging from hand-crafted features~\cite{hoiem2007recovering, liu2008sift, saxena2008make3d}, convolutional neural networks~\cite{eigen2014depth, laina2016deeper, chen2016single, xu2017multi, qi2018geonet, fu2018deep, ranftl2020robust}, to transformers~\cite{Ranftl_2021_ICCV}. The success of transformer-based techniques also prompted larger-scale data-driven approaches that mixed datasets~\cite{eftekhar2021omnidata} and utilized synthetic data to better generalize to unseen data~\cite{ke2023repurposing}. Recently, Depth Anything~\cite{yang2024depth} showed great monocular depth estimation quality by utilizing both large-scale labeled and unlabeled data.

However, most of these advances focus on natural images captured outdoors or indoors with cellphone cameras. These methods typically do not generalize well to endoscopy videos due to a lack of strong geometric features, challenging illumination conditions, and strong reflections from mucus layers. Until recently, endoscopy data with ground-truth depth maps was difficult to obtain. Thus self-supervised or unsupervised techniques utilizing limited data and techniques such as weak depth supervision~\cite{liu2019dense, luo2019details, hwang2021unsupervised} and uncertainty estimation~\cite{rodriguez2022uncertain} showed promise. Multi-view approaches~\cite{luo2019details} combined with SLAM pipelines that integrate tracking and depth estimation~\cite{ma2021rnnslam, ozyoruk2021endoslam, recasens2021endo} further showed promise despite being encumbered by weak texture cues, limited degrees of camera motion, and the presence of organ tissue deformations. Recent efforts in creating large-scale synthetic datasets of colonoscopy videos by using computer graphics engines (SimCol3D~\cite{rau2023simcol3d}) or phantom models (C3VD~\cite{Bobrow_2023}), as well as unlabeled datasets composed of clinical sequences, e.g., Colon10K~\cite{ma2021colon10k} and EndoMapper~\cite{Azagra_2023}, improved monocular depth estimation by utilizing both synthetic and real data~\cite{wang2023surfacenormal}. However, these approaches often fail to generalize to clinical data, performing extremely badly on non-axial views (comprising 40-50\% of the data), where the camera is looking away from the tubular colon structure, due to lack of strong geometric priors. In this paper, we propose a monocular depth estimation algorithm that leverages illumination information along with geometric features.

\noindent \textbf{Depth Using Illumination Information.} Researchers have developed techniques where one can reconstruct a surface from multiple images captured with varying illumination conditions, also called Photometric Stereo~\cite{basri2007photometric, logothetis2019differential, lichy2021shape, yang2022ps, zhao2023mvpsnet}. One special focus is near-field Photometric Stereo, where the light source is close to the object resulting in different points on the surface receiving different intensities of light. Numerous approaches to near-field photometric stereo utilize a per-pixel lighting representation to effectively model scene lighting and for subsequent estimation or refinement of surface normals in an optimization framework~\cite{papadhimitri2014uncalibrated, logothetis2020cnn, santo2020deep} or as an input to a neural network ~\cite{lichy2022fast}. We also use per-pixel lighting information, but go two steps forward, to create a supervised and self-supervised loss function that enables us to train on unlabeled real clinical data. Additionally, we create a depth refinement network that uses a per-pixel shading map to modulate an initial depth prediction. In contrast to previous near-field photometric stereo approaches we only require a single image captured with a co-located light and camera.

Recently, LightDepth~\cite{rodríguezpuigvert2023lightdepth} also proposed self-supervised loss functions that used illumination attenuation with differentiable rendering for image reconstruction. Our proposed self-supervised loss operates on per-pixel shading representation and avoids explicit rendering of an image with imperfect assumptiosn of diffuse albedo and direct lighting in LightDepth~\cite{rodríguezpuigvert2023lightdepth}. We also go beyond self-supervised loss, by developing supervised per-pixel shading loss, teacher-student transfer learning and illumination-aware depth refinement module, producing significantly better result than LightDepth on both synthetic and clinical data.

\noindent \textbf{Depth Refinement Network.} Depth refinement is a commonly employed technique that leverages an initial predicted depth and additional information, such as surface normals, multi-scale information, or depth context, in order to produce higher quality depth maps~\cite{kwon2015data, khamis2018stereonet, rossi2020joint}. Recent approaches utilize geometric consistency, uncertainty estimation, or guidance masks to refine depths~\cite{kusupati2020normal, bae2022irondepth, kim2022layered}. We propose a depth refinement module that modulates an initial depth prediction with extracted RGB features and features extracted from Per-Pixel Shading, which is calculated from the initial depth prediction, to generate the refined depth - a first to our knowledge.

\noindent \textbf{Teacher-Student Transfer Learning.} Teacher-student transfer learning is a popular form of knowledge distillation\cite{hinton2015distilling, gou2021knowledge,romero2014fitnets, zagoruyko2016paying, furlanello2018born}. Depth Anything~\cite{yang2024depth}, a monocular depth estimation technique, utilizes teacher-student learning while leveraging large-scale data (1.5 million labeled images, 62 million unlabeled images) and training with a self-supervised loss function that includes semantic losses. For endoscopy, such large-scale data is unavailable and semantic losses are not meaningful. Our proposed teacher-student learning uses a lighting-aware self-supervised loss function using only $\sim$ 18,000 images.

\begin{figure}[tb]
  \centering
  \includegraphics[width=\linewidth]{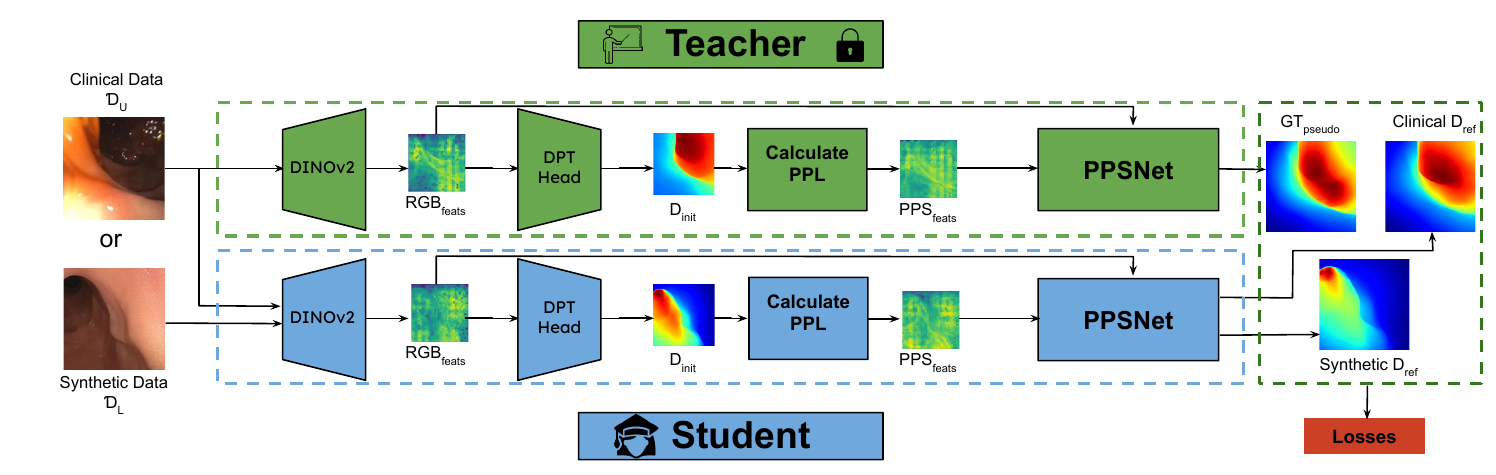}
  \caption{An overview of our proposed approach to train a student network capable of producing high quality depth maps on both synthetic colonoscopy data and real clinical colonoscopy data. For simplicity, $PPSNet$ here corresponds to lines 4-7 of alg.~\ref{alg:main_algo}.}
  \label{fig:main_arch}
  \vspace{-2em}
\end{figure}

\vspace{-1em}
\section{Our Approach}
\vspace{-1em}

Our approach utilizes per-pixel-shading (PPS) fields as a part of our learning objectives and refinement module to produce a model capable of achieving the state-of-the-art monocular depth estimation on the synthetic C3VD~\cite{Bobrow_2023} colonoscopy dataset. In the following sections, we describe the analytical computation of per-pixel lighting, the PPS field in Sec. \ref{sec:ppl}, and its subsequent use in both our learning objectives (Sec. \ref{sec:ppl-loss}) and as an input to our novel depth refinement module (Sec. \ref{sec:archi}). We then propose a teacher-student transfer learning that learns from unlabeled, real clinical data using our proposed self-supervised loss function, described in Sec. \ref{sec:tsl}.

\vspace{-1em}
\subsection{Per-Pixel Lighting Representation}
\label{sec:ppl}

Existing depth estimation techniques \cite{ranftl2020robust, Ranftl_2021_ICCV, ke2023repurposing, yang2024depth} rely on learning geometric and semantic priors from their training data. However, these methods ignore crucial photometric information that can be obtained from a near-field co-located light and camera. The surfaces of an internal organ closer to the camera and facing the camera receive more incident illumination than surfaces further from the camera or facing in opposite directions. Our key idea is to leverage this fact by modeling incident illumination at every point on the surface using a Per-Pixel Lighting (PPL) representation, following ~\cite{lichy2022fast}. However, our method improves upon Lichy \textit{at al.}~\cite{lichy2022fast}, which only uses per-pixel lighting as an input to a neural network for surface normal estimation, by explicitly designing loss functions to train on synthetic and unlabeled real data and by designing a depth refinement architecture using feature modulation. In this paper, we focus on colons due to publicly available datasets, but our method is general and can be extended to other organs like the upper airway.

\textbf{Camera Model.} We utilize the conventional pinhole camera framework, positioned at the origin in the global coordinate system and looking down the z-axis. This camera is defined by the intrinsic matrix $K$. A point in the world space, denoted as $X = (x, y, z)$, is mapped to a pixel coordinate $(u, v)$ using: $(u,v,1)^T \sim K(x,y,z)^T$.

\textbf{Geometry Model.} We only take into account visible surfaces of the colon, and therefore assume that the depth map and corresponding normal map is a complete description of in-view surfaces. $X(u,v) \in \mathbb {R}^3$ describes the 3D location of a pixel $(u,v)$ on the surface of the organ. The depth map can then be defined by $D(u,v) = X(u,v)_{3}$, corresponding to the z-component of $X(u,v)$. $X(u,v)$ itself can be recovered from the following: $X(u, v) = D(u, v)K^{-1}(u,v,1)^T$.

\textbf{Per-Pixel Lighting.} Following the notations introduced in ~\cite{lichy2022fast}, we represent any point light source, i.e. the co-located light of the endoscope, with its location $p \in \mathbb{R}^3$, light direction  $d \in \mathbb{S}^2$, and angular attenuation coefficient $\mu \in \mathbb{R}$. Then for any point $X$ on the surface of the colon, we can represent Per-Pixel Lighting (PPL) using a lighting direction $L(X)$ and an attenuation factor $A(X)$ as follows: 
\begin{equation}
L(X) = \frac{X-p}{\lVert X-p \rVert}\,,\quad A(X) = \frac{(L \cdot d)^\mu}{\lVert X-p \rVert^{2}}\,.
\label{eq:ppl_terms}
\end{equation}
Thus we can analytically compute PPL information $L(u,v)$ and $A(u,v)$ for every pixel $(u,v)$ of an image captured with a co-located light source and camera, i.e., an endoscope, given the depth map $D(u,v)$ and the camera intrinsics $K$. This presents a chicken-and-egg problem where the PPL information is strongly correlated with depth, i.e. one can be computed from the other, despite both being unknown initially. Our proposed solution to this problem is to leverage ground truth and pseudo-ground truth where possible, as well as a self-supervised loss objective that takes advantage of the aforementioned strong correlation. For our calculation of per-pixel lighting (PPL) information, we assume the camera and light source are co-located with the camera direction being in the positive z direction. Additionally, we assume that our light is modeled by an isotropic point source. Therefore, the angular attenuation coefficient $\mu$ is 0 (rendering $d$ in the calculation of $A(X)$ in eqn.~\ref{eq:ppl_terms} irrelevant).

\vspace{-0.5em}
\subsection{Proposed Loss Functions Using Per-Pixel Lighting}
\vspace{-0.5em}

\label{sec:ppl-loss}

Following the PPL representation, introduced in ~\cite{lichy2022fast} and summarized in Sec. \ref{sec:ppl}, we propose two new loss functions, a supervised loss function for training on synthetic data, and a self-supervised loss function for training on clinical data.

\begin{figure}[H]
  \centering
  \includegraphics[width=\linewidth]{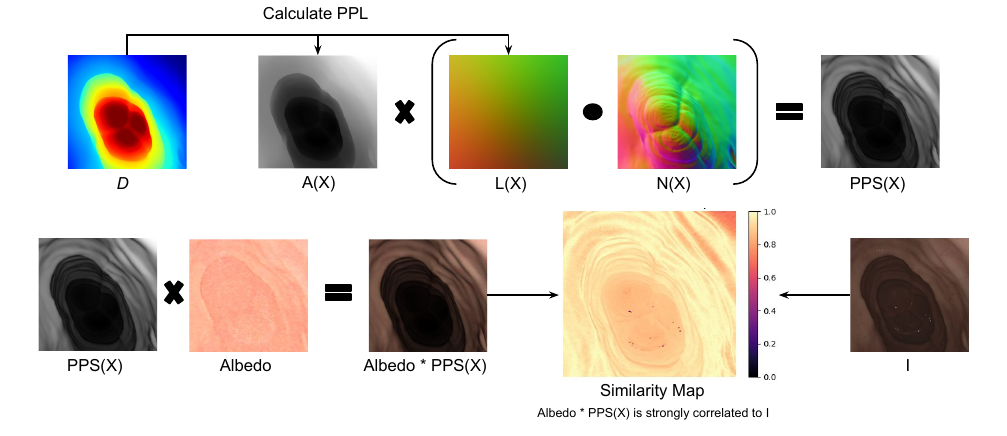}
  \vspace{-2em}
  \caption{An example of how we compute our PPS representation using depths and surface normals. The computed albedo-modulated PPS representation is strongly correlated to the corresponding input image.}
  \label{fig:ppl_vis}
\end{figure}

First, we propose a new representation, the Per-Pixel Shading field $PPS$, that combines the analytically computed PPL information of lighting direction, $L(X)$, and attenuation factor, $A(X)$, with the surface normal $N(X)$ to capture the effects of incident lighting on the surface geometry:
\begin{equation}
PPS (X) = A(X) \times (L(X) \cdot N(X)).
\label{eq:pps_field}
\end{equation}
The surface normal can be computed by differentiating the depth map $D(u,v)$. A visual illustration of our proposed PPS field can be found in Fig.~\ref{fig:ppl_vis}. PPS captures the diffuse shading effect, which indicates how light emitted from the endoscope will be reflected by the surface, ignoring strong specularities and inter-reflections.

\textbf{Supervised PPS Loss.} We introduce a loss function that computes the mean squared error between the ground-truth Per-Pixel Shading (PPS), computed using the ground truth depths and surface normals, and the predicted PPS, analytically computed from a final, refined depth estimate. For an image of resolution $H \times W$, we can compute this loss using eqn.~\ref{eq:ppl_field_loss} where specularities are masked with mask function $M$, s.t. $M(u,v) = 1, ~ if ~I_g < 0.98$, where $I_g$ is the intensity map of the input image. We believe that the supervised PPS loss is complementary to supervised depth loss due to the inherent smoothness and low-frequency bias in the depth maps compared to more high-frequency information in the PPS map. Additionally, the lighting information in the PPS map can also help to better resolve distance ambiguity using the inverse squared fall-off. Our experimental analysis in Tab. \ref{tab:teacher_ablation} also shows that on the synthetic C3VD dataset using $\mathcal{L}_{PPS-sup}$ helps to improve depth prediction.
\begin{equation}
\mathcal{L}_{PPS-sup} = \frac{1}{HW} \sum_{u=1}^{H} \sum_{v=1}^{W} M(u,v) \left( PPS_{pred}(u,v) - PPS_{gt}(u,v) \right)^2.
\label{eq:ppl_field_loss}
\end{equation}

\textbf{Self-supervised PPS Loss.} We introduce an additional loss function that utilizes the Per-Pixel Shading (PPS) representation but does not require the knowledge of ground-truth depths and surface normals and thus can be used for self-supervised learning. Our idea relies on the fact that the PPS is strongly correlated with the image intensity field except in regions of strong specularity and it ignores inter-reflections by only modeling direct, in-view illumination from the surface to the camera. We show this correlation in Fig. \ref{fig:ppl_vis}, and we observe that on the synthetic C3VD dataset that utilizes a phantom colon mold, the ground-truth PPS has a mean correlation of 0.90 with a variance 0.03 with the image intensity field. This high correlation indicates that the model predictions for PPS are closely aligned with changes in the image intensity field, suggesting consistent and accurate modeling. The low variance points towards the stability of this relationship across different samples, meaning that our approach's use of the PPS is uniform and dependable across the entire dataset.

\vspace{-2em}
\vspace{-2em}

Based on this observation, we introduce a self-supervised loss function that computes the correlation between the PPS, derived from the predicted depth map, and the grayscale image (intensity field) \(I_{g}\), where specularities are masked with mask function $M$ using a threshold of 0.98:
\vspace{-1em}

\begin{equation}
\mathcal{L}_{PPS-corr} = 1 - \text{corr}(M*I_{g}, M*PPS_{pred}).
\label{eq:ppl_corr_loss}
\end{equation}

The self-supervised loss function will enable us to train on real clinical data where ground-truth depth information is unavailable. Further details regarding the calculation of $\mathcal{L}_{PPS-corr}$ can be found in our supplementary materials.

\vspace{-1em}
\subsection{Depth Refinement Module}
\vspace{-0.5em}
\label{sec:archi}

We propose to use illumination information, represented by Per-Pixel Shading to enhance the depth refinement process. Our depth refinement module, called PPSNet, takes in RGB features ($RGB_{feats}$), $PPS$ features ($PPS_{feats}$), and an initial depth prediction ($D_{init}$) in order to predict a value $\Delta_{D}$ to refine $D_{init}$ with. All features are extracted using a DINOv2~\cite{oquab2024dinov2} encoder. The initial depth prediction is obtained from a DPT-based decoder as described in~\cite{yang2024depth}. Since DINOv2 expects an RGB image, we compute PPS features by simply multiplying the calculated Per-Pixel Shading (PPS) map with a proxy albedo, effectively an $H, S$ value pair extracted from the input RGB image with $V$ set to 100\%. Due to the colon surfaces having a small variance in albedo, it is reasonable to create and use such a proxy for albedo. Integration of these features is achieved through a multi-headed cross-attention ~\cite{vaswani2017attention}, which can be simply viewed as eqn.~\ref{eq:cross_attention}:
\begin{equation}
x_{combo} = \text{CrossAttention}(Q, K) = \text{Softmax}\left(\frac{QK^T}{\sqrt{d_k}}\right)V,
\label{eq:cross_attention}
\end{equation}
where $Q = RGB_{feats}$, $K = V = PPS_{feats}$, $d_k$ denotes the dimensionality of the keys, and $x_{combo}$ represents the synthesized feature set. These combined features influence $D_{init}$ via feature-wise linear modulation (FiLM~\cite{perez2017film}), altering it to $D_{mod}$ in eqn.~\ref{eq:film}:
\begin{equation}
D_{mod} = \gamma(x_{combo}) \cdot D_{init} + \beta(x_{combo}),
\label{eq:film}
\end{equation}
where $\gamma$ and $\beta$ are the scale and shift parameters generated by the FiLM module, dependent on $x_{combo}$. 

Following modulation, $D_{mod}$ is fed into a four-layer UNet, producing $\Delta_{D}$, which is then finally added to $D_{init}$ to yield $D_{ref}$. This sequence of steps describes our approach to using PPS for refinement and is further detailed within algorithm~\ref{alg:main_algo}, which takes in the input image $I$, light position $p$, light direction $d$, angular attenuation coefficient $\mu$, and camera intrinsics $K$. Additional insights into our approach are available in the supplementary materials.

\begin{figure}[H] 
\vspace{-4em}
\begin{algorithm}[H]
\caption{A forward pass of our approach including PPSNet (lines 4-7)}\label{alg:main_algo}
\begin{algorithmic}[1] 
\Require $I$, $p$, $d$, $u$, $K$
\State $RGB_{feats} = DINOv2\_Encoder(I)$
\State $D_{init} = DPT\_Decoder(RGB_{feats})$
\State $PPS_{feats} = CalculatePPL(D_{init}, p, d, u, K)$ \Comment {involves eqn.~\ref{eq:ppl_terms}}
\State $x_{combo} = CrossAttention(RGB_{feats}, PPS_{feats})$ \Comment {involves eqn.~\ref{eq:cross_attention}}
\State $D_{mod} = FiLM\_Mod(D_{init}, x_{combo})$ \Comment {involves eqn.~\ref{eq:film}}
\State $\Delta_{D} = UNet(D_{mod})$
\State $D_{ref} = D_{init} + \Delta_{D}$
\end{algorithmic}
\end{algorithm}
\label{fig:main_algo} 
\vspace{-4em}
\end{figure}

\subsection{Teacher-Student Transfer Learning for Sim2Real}
\vspace{-0.5em}

\label{sec:tsl}

\noindent \textbf{Teacher network.} We first train our proposed network architecture that combines a foundational monocular depth estimation network followed by PPSNet, as shown in algorithm \ref{alg:main_algo} on synthetic C3VD~\cite{Bobrow_2023} data with supervision. Specifically, we train with scale-shift invariant ($\mathcal{L}_{SSI}$), depth regularization ($\mathcal{L}_{REG}$), and virtual-normal ($\mathcal{L}_{VNL}$) losses described in prior works~\cite{ranftl2020robust, eftekhar2021omnidata}, along with our proposed $\mathcal{L}_{PPS-sup}$ loss. While this network produces state-of-the-art performance on the C3VD dataset we observe that it often produces errors and artifacts on real clinical data. This is a common problem for any neural network trained on synthetic data and tested on real data due to the Sim2Real gap. 

To address this problem we propose Teacher-Student transfer learning, where we train a student network on real data with self-supervision and the guidance of the Teacher network.  A similar approach is also utilized by a recent state-of-the-art monocular depth estimation algorithm~\cite{yang2024depth} that leverages large-scale data (1.5 million labeled and 62 million unlabeled images) and trains using a self-supervised loss approach with semantic losses. Such a large-scale unlabeled data is unavailable for endoscopy and semantic losses are meaningless for colons. Instead, we propose a teacher-student transfer learning that utilizes illumination information to train on $\sim$18,000 real clinical images~\cite{ma2021colon10k, wang2024structurepreservingimagetranslationdepth} using self-supervision.

\noindent \textbf{Student network.}  Our Student network has the same architecture as the Teacher network. The Student model is trained using both labeled synthetic data ($\mathcal{D}_{L}$) and unlabeled clinical data ($\mathcal{D}_{U}$) in an interleaved manner, which provides more stability during training. When using labeled synthetic data we use the same supervised loss functions as the teacher. However, when training on unlabeled real data we have no supervision w.r.t. ground-truth depth and normal map. Instead, we use the proposed self-supervised loss function $\mathcal{L}_{PPS-corr}$ for training. However, self-supervision on its own is not powerful enough to produce accurate depth estimates due to inherent ambiguity in training. Thus we use guidance from the Teacher network in terms of pseudo-supervision, where the output of the Teacher network for an input clinical image is used as a pseudo ground-truth for computing the supervised loss functions over depth and PPS. Experimental analysis in Figure~\ref{fig:clinical_results} shows how teacher-student transfer learning improves depth estimation quality on real clinical data. Further details regarding our teacher-student learning approach, including a corresponding algorithm table, can be found in our supplementary materials.

\vspace{-1em}
\subsection{Implementation details}
\label{sec:implementation}

We implement our approach in PyTorch~\cite{paszke2019pytorch} and use the AdamW~\cite{kingma2014adam, loshchilov2017decoupled} optimizer coupled with the OneCycleLR policy~\cite{smith2019super} for training. For the OneCycleLR policy, we choose a maximum learning rate of 1e-5 and train for 20 epochs with a batch size of 8 on 4 x NVIDIA RTX A6000s (48GB each). We also utilize the initial weights from~\cite{yang2024depth} as a starting point for all of our baselines and results in order to leverage the rich, prior knowledge of state-of-the-art monocular depth estimation methods. It takes approximately 5 hours to train our teacher model and approximately 10 hours to train our student model.

\vspace{-1em}
\section{Results}
\label{sec:results}
\vspace{-1em}

\subsection{Experimental Setup}
\vspace{-0.5em}

\subsubsection{Datasets.} We train and test on separate splits of the C3VD~\cite{Bobrow_2023} dataset and our clinical dataset composed of oblique and en face views~\cite{wang2024structurepreservingimagetranslationdepth} and Colon10K~\cite{ma2021colon10k}. C3VD includes 22 video sequences amounting to 10,015 frames with comprehensive ground truth annotations. Our clinical dataset includes 80 sequences with oblique views (7,293 frames), 14 sequences with en-face views (832 frames), and 20 sequences with down-the-barrel (axial) views (10,216 frames). For training and testing with C3VD, we re-use the splits noted in the supplementary materials of LightDepth~\cite{rodríguezpuigvert2023lightdepth}. Further details are provided in our supplementary.

\vspace{-1em}
\subsubsection{Baselines.} We compare our approach to the state-of-the-art monocular depth estimation techniques on endoscopy videos. $\bullet$ UNet~\cite{ronneberger2015u} is a meaningful baseline when provided with ground truth depth due to its ability to capture multi-scale contextual information and precisely localize features through its symmetric expanding path. $\bullet$ MonoDepth2~\cite{godard2019digging} and NormDepth~\cite{wang2023surfacenormal} are promising, self-supervised approaches that leverage multi-view photometric consistency to enforce geometric constraints across different viewpoints. $\bullet$ We also compare with \textit{Wang et al.}~\cite{wang2023surfacenormal} which includes NormDepth as a part of a SLAM pipeline for joint normal and depth refinement. $\bullet$ LightDepth~\cite{rodríguezpuigvert2023lightdepth} is a recent approach that is completely self-supervised and leverages the principle of illumination decline alongside a photometric model. $\bullet$ We also compare to the current state-of-the-art in monocular depth estimation, Depth Anything~\cite{yang2024depth}, which is not trained on endoscopy images. $\bullet$ We then present Ours-Backbone, the backbone model that fine-tunes a pre-trained DINOv2~\cite{oquab2024dinov2} encoder (small) and DPT~\cite{Ranftl_2021_ICCV} decoder used in~\cite{yang2024depth} using supervised loss functions on the depth map, without any PPS based losses. $\bullet$ We then present Ours-Teacher network, which combines Ours-Backbone with the PPSNet depth refinement module and trains on C3VD with supervised depth and PPS losses. $\bullet$ Finally, Ours-Student model, has the same architecture as Ours-Teacher but is trained on both C3VD and real clinical data using supervised and self-supervised PPS loss functions respectively.

\vspace{-1em}
\subsubsection{Metrics.} 
\label{sec:metrics} We evaluate commonly used depth metrics such as root-mean-squared error ($RMSE$), absolute relative error ($AbsRel$), squared relative error ($SqRel$) multiplied by a factor of 1000, and $\delta < 1.1$ which indicates percentage of pixels within 10\% of the actual depth values. 

\begin{table}[tb]
\centering
\caption{We compare our approach with existing monocular depth estimation techniques developed for endoscopy videos by testing on the synthetic C3VD dataset. Ours-Student trains using both C3VD and real data. Best results are shown in bold. Second best results are underlined.}
\resizebox{\textwidth}{!}{
\begin{tabular}{lccccccc}
\toprule
Architecture & Encoder & Supervision & $RMSE$ $\downarrow$ & $AbsRel$ $\downarrow$ & $SqRel$ $\downarrow$ & $\delta < 1.1$ $\uparrow$ \\
\midrule
UNet~\cite{ronneberger2015u} & ResNet18 & GT & 6.71 & 0.126 & 1.72 & 0.63 \\
MonoDepth2~\cite{godard2019digging} & ResNet50 & SSL & 9.11 & 0.192 & 2.64 & 0.43 \\
NormDepth~\cite{wang2023surfacenormal} & ResNet50 & SSL & 9.72 & 0.164 & 2.05 & 0.42 \\
Wang et al.~\cite{wang2023surfacenormal} & ResNet50 & GT \& SSL & 7.51 & 0.155 & 1.53 & 0.48 \\
LightDepth DPT~\cite{rodríguezpuigvert2023lightdepth} & DPT-Hybrid & SSL & 6.55 & 0.0780 & 1.81 & 0.56 \\
DepthAnything~\cite{yang2024depth} & DINOv2 & GT \& SSL & 12.76 & 0.283 & 6.41 & 0.34 \\
\midrule
Ours - Backbone & DPT-Hybrid & GT & 3.31 & 0.0755 & 1.66 & 0.72 \\
Ours - Backbone & DINOv2 & GT & 2.79 & 0.0622 & 0.238 & 0.82 \\
Ours - Teacher & DPT-Hybrid & GT & 2.68 & 0.0716 & 0.97 & 0.83 \\
Ours - Teacher & DINOv2 & GT & \underline{2.15} & \underline{0.0529} & \underline{0.147} & \underline{0.87} \\
Ours - Student & DINOv2 & GT \& SSL & \textbf{2.06} & \textbf{0.0491} & \textbf{0.140} & \textbf{0.89} \\
\bottomrule
\end{tabular}}
\label{tab:sota_results}
\vspace{-1em}
\end{table}

\vspace{-0.5em}
\subsection{Comparison with Existing Approaches}
\label{sec:ppsnet}

\textbf{Quantitative Evaluation.} In Table~\ref{tab:sota_results}, we compare to the state-of-the-art approaches on the C3VD~\cite{Bobrow_2023} dataset that has paired ground truth depth. We include both Ours-Teacher model (trained only on C3VD data) and Ours-Student model (trained on both C3VD data and clinical data) for comparison. Unless indicated otherwise, all other baselines were trained only on C3VD data. 

\subsubsection{Qualitative Results.} 
\begin{figure}[tb]
  \centering
  \vspace{-1em}
  \includegraphics[width=\linewidth]{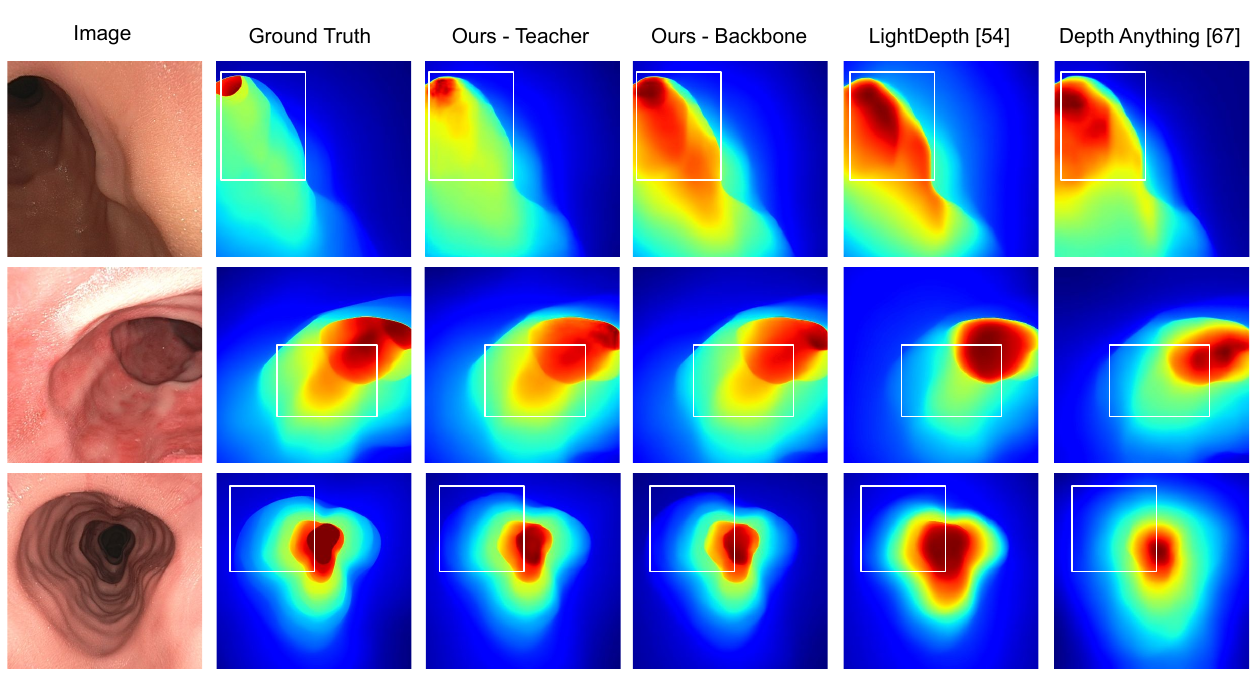}
  \caption{Qualitative evaluation on the C3VD dataset. Red = further distance from the camera and blue is closer. Regions to note during visual comparison are outlined in white. Ours-Teacher performs best, improving upon Ours-Bakcbone due to our proposed depth refinement PPSNet and self-supervised PPS loss.}
  \label{fig:c3vd_qualitative_results}
  \vspace{-2em}
\end{figure}

We show qualitative results in the form of relative depth maps for both the C3VD dataset and our clinical dataset composed of down-the-barrel (axial), oblique, and en-face views. As shown in Figs. \ref{fig:c3vd_qualitative_results} and \ref{fig:clinical_results}, our method produces higher-quality depth maps with fewer discrepancies in the case of C3VD and clinical data.

\subsubsection{Ablation Studies.}
\label{sec:teacher_ablation}

\vspace{-2em}
\noindent \textbf{Role of PPSNet and $\mathcal{L}_{PPS-sup}$ in improving Ours-Teacher over Ours-Backbone.} 

\begin{wraptable}[10]{r}{0.5\textwidth}
\vspace{-3em}
\centering
\caption{We show that both PPSNet depth refinement and $\mathcal{L}_{PPS-sup}$ loss helps in improving the performance of Ours-Teacher (row 4) over Ours-Backbone (row 1).}
\begin{tabular}{ccccc}
\toprule
$\mathcal{L}_{PPS-sup}$ & PPSNet & RMSE $\downarrow$ & AbsRel $\downarrow$ \\
\midrule
\xmark & \xmark & 2.79 & 0.0621 \\
\cmark & \xmark & 2.21 & 0.0518 \\
\xmark & \cmark & 2.60 & 0.0617 \\
\cmark & \cmark & 2.15 & 0.0529 \\
\bottomrule
\end{tabular}
\label{tab:teacher_ablation}
\vspace{-2em}
\end{wraptable}

\begin{figure}[H]
  \centering
  \includegraphics[width=\linewidth, height=0.7\textheight, keepaspectratio]{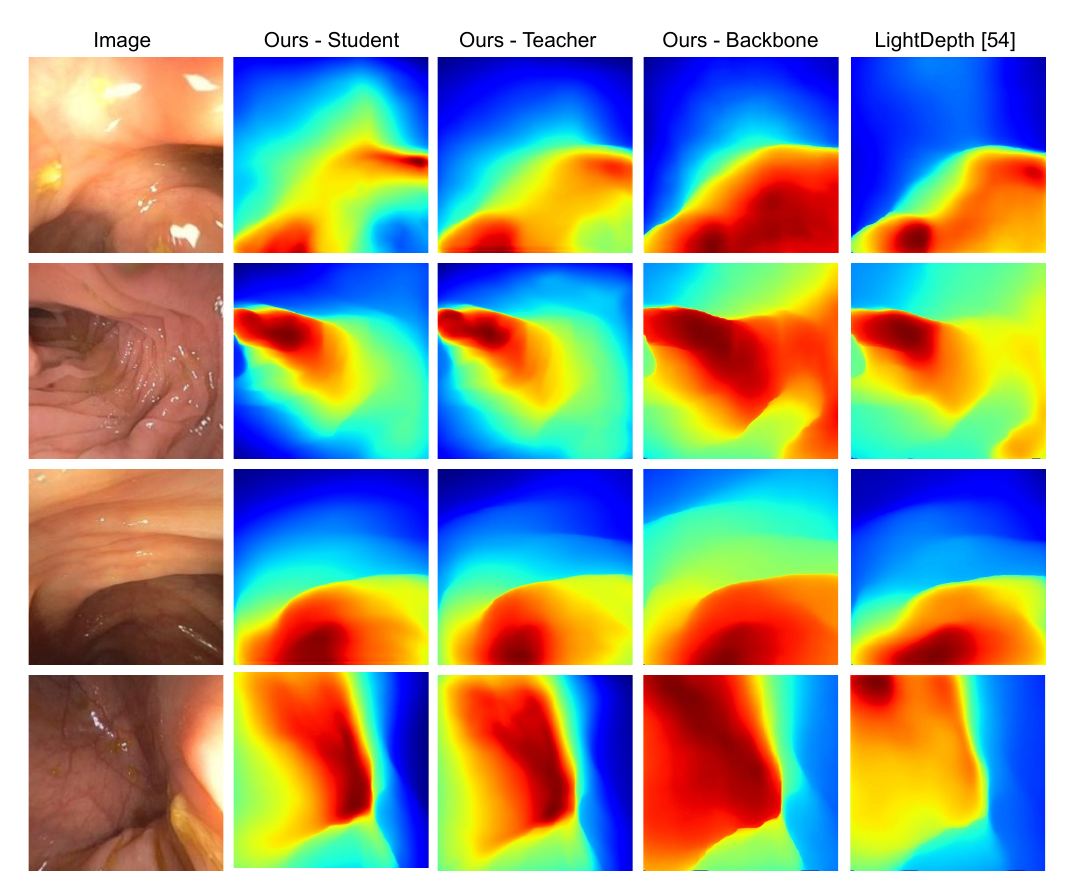}
  \caption{Qualitative evaluation on clinical data. Red = further distance from the camera and blue is closer. Additional results can be found in the appendices.}
  \label{fig:clinical_results}
\end{figure}

In Table \ref{tab:sota_results} and Figure \ref{fig:c3vd_qualitative_results} and \ref{fig:clinical_results} we showed that Ours-Teacher is better than Ours-Backbone. We propose two distinct modifications over Ours-Backbone, involving depth refinement module PPSNet and supervised loss function $\mathcal{L}_{PPS-sup}$. In Table \ref{tab:teacher_ablation} we show that both PPSNet and $\mathcal{L}_{PPS-sup}$ help in improving the performance of Ours-Teacher (row 4) over Ours-Backbone (row 1).

\noindent \textbf{Role of features extracted from PPS representation in the depth refinement network PPSNet.} In Table \ref{tab:refinement_module_ablation} we show that modulating initial depth features extracted from input images using DINOv2 with RGB features ($RGB_{feats}$) and PPS features ($PPS_{feats}$) improves the depth refinement performance.

\noindent \textbf{Role of teacher-student learning.} In Table \ref{tab:sota_results} and Figure \ref{fig:clinical_results} we showed that Ours-Student improves over Ours-Teacher, noticeably on real clinical data since Ours-Teacher is only trained on C3VD and hence fails to generalize well on real clinical data. The strong performance of Ours-Student can be attributed to two key contributions. Firstly we show in Table~\ref{tab:teacher_student_ablation} that using self-supervised PPS loss function $\mathcal{L}_{PPS-corr}$ on real data helps in improving the performance, even when testing on synthetic data. 

\begin{wraptable}[16]{r}{0.5\textwidth}
\centering
\caption{We show that $RGB_{feats}$ and $PPS_{feats}$ improves depth refinement on C3VD when used for modulation.}
\begin{tabular}{cccccc}
\toprule
$RGB_{feats}$ & $PPS_{feats}$ & RMSE $\downarrow$ & AbsRel $\downarrow$ \\
\midrule
\cmark & \xmark & 2.29 & 0.0531 \\
\xmark & \cmark & 2.24 & 0.0551 \\
\cmark & \cmark & 2.15 & 0.0529 \\
\bottomrule
\end{tabular}
\label{tab:refinement_module_ablation}

\end{wraptable}

Additionally, we observe that having a strong teacher module trained on synthetic data provides a general prior for colon geometry and has a measurable benefit over just using DINOv2 features from a teacher network. Finally, we note that the most optimal performance across all metrics is obtained by using both our proposed teacher and $\mathcal{L}_{PPS-corr}$. We also show in~\autoref{tab:rebuttal_impact_of_PPS} that the impact of our PPS representation in particular as a part of loss objectives is still superior across both `Ours - Teacher` and `Ours - Student` in contrast to LightDepth~\cite{rodríguezpuigvert2023lightdepth} and its losses.

\begin{wraptable}{r}{0.5\textwidth}
\vspace{-12em}
\centering
\caption{We show that both self-supervised $\mathcal{L}_{PPS-corr}$ and a strong teacher network Ours-Teacher helps in improving the performance of Ours-Student.}
\begin{tabular}{cccccc}
\toprule
Teacher & $\mathcal{L}_{PPS-corr}$ & RMSE $\downarrow$ & AbsRel $\downarrow$ \\
\midrule
DINOv2 & \xmark & 2.55 & 0.0625 \\
DINOv2 & \cmark & 2.31 & 0.0533 \\
Ours & \xmark & 2.28 & 0.0544 \\
Ours & \cmark & 2.06 & 0.0491 \\
\bottomrule
\end{tabular}
\label{tab:teacher_student_ablation}
\vspace{-2em}
\end{wraptable}

\noindent \textbf{Role of encoder (backbone) and losses.} It is clear that a variety of components (e.g., backbones, loss strategies) can have significant effects that can cast doubts on the results shown in~\autoref{tab:sota_results}. To help alleviate these doubts, we note the encoder (backbone) differences in~\autoref{tab:sota_results} and an ablation of proposed losses in~\autoref{tab:rebuttal_impact_of_PPS} showing that the impact of our approach is significant regardless of the encoder (backbone) utilized and in contrast to LightDepth's~\cite{rodríguezpuigvert2023lightdepth} rendering-based approach.

\begin{wraptable}{r}{0.4\textwidth}
\vspace{-4em}
\centering
\caption{Our proposed losses involving PPS representation are superior to LightDepth (LD) rendering~\cite{rodríguezpuigvert2023lightdepth}.}
\begin{tabular}{cccc}
\toprule
Model & Loss & RMSE $\downarrow$ & AbsRel $\downarrow$ \\
\midrule
Teacher  & LD & 2.44 & 0.0588 \\
Teacher & Ours & \underline{2.15} & \underline{0.0529} \\
Student & LD & 2.29 & 0.0546 \\ Student & Ours & \textbf{2.06} & \textbf{0.0491} \\
\bottomrule
\end{tabular}
\label{tab:rebuttal_impact_of_PPS}
\vspace{-2em}
\end{wraptable}



\section{Limitations and Discussion}
\vspace{-1em}

Despite the success of our approach and its ability to be extended to any monocular depth estimation method through the adoption of our learning objectives and our refinement module, we note that the formulation of per-pixel lighting that we utilize does not adequately model albedo and does not directly address in-view, specular lighting effects that can lead to misinterpreted lighting cues and ultimately inaccurate depth estimations. To remedy this, we opt to detect and mask out specular lighting effects. Future works should aim to model such effects and evaluate their impact on downstream tasks such as 3D mesh reconstruction. Furthermore, we acknowledge that the main benefit of our approach still comes from depth supervision. We believe this means that future works further exploring self-supervision as~\cite{rodríguezpuigvert2023lightdepth} did have potential promise, especially with unlabeled, clinical data. 
\clearpage

\section{Conclusion}
\vspace{-0.75em}

This paper presents a novel approach to leveraging per-pixel lighting (PPL) information toward enhanced depth estimation in colonoscopy. We use supervised and self-supervised variants of loss functions that use an analytically computed per-pixel shading (PPS) field, which is also used as an input into our refinement network that conditions an initial predicted depth on a combination of RGB features and PPS features. Our approach achieves state-of-the-art results on the C3VD dataset while enabling the estimation of high-quality depth maps on real clinical data. We do not feel that our approach is limited to colonoscopy, nor bio-applications such as endoscopy for that matter. Our approach can be useful in other bio-applications such as bronchoscopy, which we demonstrate in our supplementary materials, as well as more mainstream contexts for depth estimation (e.g., in egocentric vision). 

\section*{Acknowledgements}

This work is supported by an NIH project (\#1R21EB035832).

\clearpage

\appendix

\section{Overview of Appendices}

Our appendices contain the following additional details and results:

\begin{itemize} 
    \itemsep0em
    \item In Sec.~\ref{sec:supp_pc}, we present point cloud results from our approach. We also provide point cloud results from LightDepth~\cite{rodríguezpuigvert2023lightdepth} for comparison.
    \item In Sec.~\ref{sec:supp_methods}, we provide additional information regarding our approach. Sec.~\ref{sec:supp_ppl_light_rep} includes additional details about the proposed per-pixel lighting (PPL) representation and its usage in our proposed approach. In Sec.~\ref{sec:supp_ts_algo}, we provide additional details regarding our approach to teacher-student learning, including a full algorithm table. Sec.~\ref{sec:supp_implementation} contains additional details regarding our implementation, including our chosen loss weights and backbone model size.
    \item Sec.~\ref{sec:supp_experiments} includes additional information regarding our experiments, chiefly the training splits we utilized that match~\cite{rodríguezpuigvert2023lightdepth}, as well as additional qualitative results.
    \item Sec.~\ref{sec:supp_code} describes our limited code release included as a part of these supplementary materials. The code includes our proposed loss functions and model files for reference, as well as pre-trained models. We will release our full code, including the training code and various baselines, in the near future.
    \item Sec.~\ref{sec:supp_lightdepth} includes additional information regarding our implementation of LightDepth~\cite{rodríguezpuigvert2023lightdepth}.
    \item We provide additional, qualitative results on bronchoscopy data in Sec.~\ref{sec:supp_bronchoscopy}.
    \item Sec.~\ref{sec:supp_clinical_dataset} includes additional details regarding the clinical dataset utilized in our work and to be released in the near future.
\end{itemize}

\clearpage

\section{Point Clouds}
\label{sec:supp_pc}
\begin{figure}[H]
\vspace{-1.5em}
  \centering
  \includegraphics[width=\linewidth]{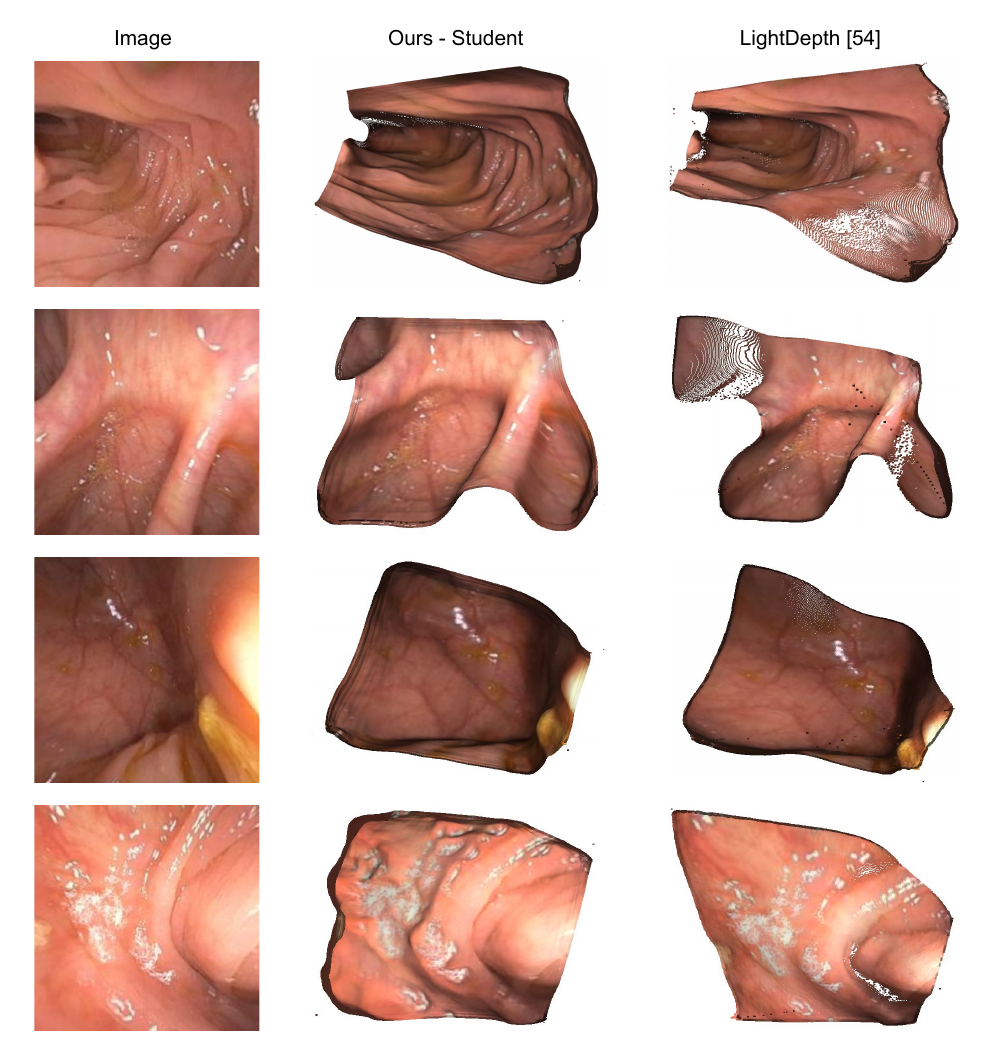}
  \caption{A qualitative view of 3D point clouds generated by our approach and in contrast to LightDepth~\cite{rodríguezpuigvert2023lightdepth}. We also provide the raw \emph{.ply} files for each point cloud as a part of our supplementary materials.}
  \label{supp_fig:pc_results}
\end{figure}

\vspace{-4em}
We present point cloud results from our monocular depth estimation approach in order to show that our produced depths can eventually be used for the task of 3D mesh reconstruction and subsequent 3D analysis. In Fig.~\ref{supp_fig:pc_results} it can be seen that our produced point clouds are superior to the point clouds produced by our re-implementation of LightDepth~\cite{rodríguezpuigvert2023lightdepth}. We also provide the raw \emph{.ply} files for each point cloud in Fig.~\ref{supp_fig:pc_results} as a part of our supplementary materials.

\section{Additional Info on Methods}
\label{sec:supp_methods}

\subsection{Per-Pixel Lighting Representation and Losses}
\label{sec:supp_ppl_light_rep}

In our main paper, we noted that we utilize the conventional pinhole camera framework, positioned at the origin in the global coordinate system and looking down the z-axis. This camera is defined by the intrinsic matrix $K$. A point in the world space, denoted as $X = (x, y, z)$, is mapped to a pixel coordinate $(u, v)$ using:
\begin{equation}
(u,v,1)^T \sim K(x,y,z)^T\,.
\label{supp_eq:camera_model}
\end{equation}

Subsequently, we noted that we only take into account visible surfaces of the colon, and therefore assume that the depth map and corresponding normal map is a complete description of in-view surfaces. $X(u,v) \in \mathbb {R}^3$ describes the 3D location of a pixel $(u,v)$ on the surface of the organ. The depth map can then be defined by $D(u,v) = X(u,v)_{3}$, corresponding to the z-component of $X(u,v)$. $X(u,v)$ itself can be recovered from the following: 

\begin{equation}
X(u, v) = D(u, v)K^{-1}(u,v,1)^T\,.
\label{supp_eq:recover_X_from_D}
\end{equation}

For completeness, we now also note that given $n(X)$ is the normal at the point $X$, then the normal map can be defined by $N(u,v) = n(X(u,v))$. $N$ is subsequently computed based on the following:

\begin{equation}
N = \frac{\frac{\partial X}{\partial u} \times \frac{\partial X}{\partial v}}{\left\|\frac{\partial X}{\partial u} \times \frac{\partial X}{\partial v}\right\|}\,.
\label{supp_eq:calculate_normal}
\end{equation}

Additionally, we note that our learning objective $\mathcal{L}_{PPS-corr}$ can be formulated as follows:

\begin{equation}
\mathcal{L}_{PPS-corr} = 1 - \frac{\sum_{h=1}^{H} \sum_{w=1}^{W} (I_{gray_{hw}} - \bar{I})(PPL_{F_{hw}} - \overline{PPL}_F)}{\sqrt{\sum_{h=1}^{H} \sum_{w=1}^{W} (I_{gray_{hw}} - \bar{I})^2 \sum_{hw} (PPL_{F_{hw}} - \overline{PPL}_F)^2}}
\label{supp_eq:ppl_corr_loss}
\end{equation}

The simplified version of this formulation is presented in our main paper as follows:

\begin{equation}
\mathcal{L}_{PPS-corr} = 1 - \text{corr}(I_{gray}, PPL_{F})
\label{supp_eq:ppl_corr_loss_simplified}
\end{equation}

As noted in the main paper, the self-supervised loss function variant will enable us to train on real clinical data where ground-truth depth information is unavailable. Code implementations of both our supervised and self-supervised loss function variants are included as a part of these supplementary materials.

\subsection{Teacher-Student Transfer Learning for Sim2Real}
\label{sec:supp_ts_algo}

Our approach with teacher-student training on labeled data $\mathcal{D}_{L}$ and unlabeled data $\mathcal{D}_{U}$ can be summarized in algo.~\ref{supp_alg:teacher_student_learning} as follows:

\begin{algorithm}
\caption{Teacher-Student Learning for Endoscopic Data}
\label{supp_alg:teacher_student_learning}
\begin{algorithmic}[1]
\Require labeled synthetic dataset: $\mathcal{D}_{L}$, unlabeled real dataset: $\mathcal{D}_{U}$
\State Initialize teacher and student models with identical architectures:
\Statex \hspace{\algorithmicindent} $M_{T}$ (Teacher), $M_{S}$ (Student)  $ \gets InitializeModels()$
\State Train $M_{T}$ on $\mathcal{D}_{L}$: $M_{T} \gets Train(M_{T}, \mathcal{D}_{L})$ 
\State ~~~~ $\mathcal{L} = \alpha_{SSI} \mathcal{L}_{SSI} + \alpha_{REG} \mathcal{L}_{REG} + \alpha_{VNL} \mathcal{L}_{VNL} + \alpha_{PPS-sup} \mathcal{L}_{PPS-sup}$
\State Freeze $M_{T}$ to prevent further updates
\State Prepare mixed dataset $\mathcal{D}_{M}$ combining $\mathcal{D}_L$ and $\mathcal{D}_U$.

\For{each batch $b$ in $\mathcal{D}_{M}$}
    \If{$b$ is from $\mathcal{D}_L$}
        \State Train $M_{S}$ on $\mathcal{D}_{L}$: $M_{S} \gets Train(M_{S}, \mathcal{D}_{L})$
        \State ~~~~ $\mathcal{L} = \alpha_{SSI} \mathcal{L}_{SSI} + \alpha_{REG} \mathcal{L}_{REG} + \alpha_{VNL} \mathcal{L}_{VNL} + \alpha_{PPS-sup} \mathcal{L}_{PPS-sup}$
    \ElsIf{$b$ is from $\mathcal{D}_U$}
        \State Train $M_{S}$ on $\mathcal{D}_{U}$: $M_{S} \gets Train(M_{S}, \mathcal{D}_{U})$
        \State ~~~~ $\mathcal{L} = \alpha_{SSI} \mathcal{L}_{SSI*} + \alpha_{REG} \mathcal{L}_{REG*} + \alpha_{VNL} \mathcal{L}_{VNL*} + \alpha_{PPS-corr} \mathcal{L}_{PPS-corr}$
        \State ~~~~ Compute $\mathcal{L}_{SSI*}$, $\mathcal{L}_{REG*}$, and $\mathcal{L}_{VNL*}$ using pseudo-supervision from $M_T$.
    \EndIf
\EndFor

\end{algorithmic}
\end{algorithm}

Where the losses used alongside our proposed losses $\mathcal{L}_{PPS-sup}$ and $\mathcal{L}_{PPS-corr}$ correspond to the scale-shift invariant ($\mathcal{L}_{SSI}$), regularization ($\mathcal{L}_{REG}$), and virtual-normal ($\mathcal{L}_{VNL}$) losses described in prior works~\cite{ranftl2020robust, eftekhar2021omnidata}.

\subsection{Implementation}
\label{sec:supp_implementation}

As a part of training our approach, we use the following $\alpha$ values that describe loss weights as shown in algo.~\ref{supp_alg:teacher_student_learning}: $\alpha_{SSI} = 1.0$, $\alpha_{REG} = 0.1$, $\alpha_{VNL} = 10.0$, $\alpha_{PPS-sup} = 0.1$, and $\alpha_{PPS-corr} = 1.0$. We utilize the ViT-Small version of the backbone architecture that's also utilized in Depth Anything~\cite{yang2024depth}.
\section{Experiments}
\label{sec:supp_experiments}

\subsection{Dataset Splits}
\label{sec:supp_experiments_dataset_splits}

We utilized the same dataset splits as LightDepth~\cite{rodríguezpuigvert2023lightdepth} and include the exact sequences used for the C3VD~\cite{Bobrow_2023} dataset in Tab.~\ref{tab:dataset_splits}. Our clinical dataset includes 80 sequences with oblique views (7,293 frames), 14 sequences with en-face views (832 frames), and 20 sequences with down-the-barrel (axial) views (10,216 frames). The oblique views and en-face views are from a to-be-released dataset described in Sec.~\ref{sec:supp_clinical_dataset}. The down-the-barrel (axial) views are from the Colon10K~\cite{ma2021colon10k} dataset. The exact clinical sequences used for training and testing are included as \emph{train.txt} and \emph{val.txt} files in our supplementary materials folder.

\begin{table}[ht]
\centering
\caption{Dataset Splits for C3VD~\cite{Bobrow_2023}}
\label{tab:dataset_splits}
\footnotesize 
\setlength{\tabcolsep}{4pt} 
\renewcommand{\arraystretch}{1.2} 
\begin{tabular}{@{}cccccc@{}}
\toprule
\textbf{Sequence} & \textbf{Texture} & \textbf{Video} & \textbf{Frames} & \textbf{Set} \\
\midrule
Cecum & 1 & b & 765 & Train \\
Cecum & 2 & b & 1120 & Train \\
Cecum & 2 & c & 595 & Train \\
Cecum & 4 & a & 465 & Train \\
Cecum & 4 & b & 425 & Train \\
Sigmoid Colon & 1 & a & 800 & Train \\
Sigmoid Colon & 2 & a & 513 & Train \\
Sigmoid Colon & 3 & b & 536 & Train \\
Transcending & 1 & a & 61 & Train \\
Transcending & 1 & b & 700 & Train \\
Transcending & 2 & b & 102 & Train \\
Transcending & 2 & c & 235 & Train \\
Transcending & 3 & b & 214 & Train \\
Transcending & 4 & b & 595 & Train \\
Descending Down & 4 & a & 74 & Train \\
\midrule
Cecum & 1 & a & 276 & Test \\
Cecum & 2 & a & 370 & Test \\
Cecum & 3 & a & 730 & Test \\
Sigmoid & 3 & a & 610 & Test \\
Transcending & 2 & a & 194 & Test \\
Transcending & 3 & a & 250 & Test \\
Transcending & 4 & a & 384 & Test \\
Descending Up & 4 & a & 74 & Test \\
\bottomrule
\end{tabular}
\end{table}

\clearpage

\subsection{Qualitative Results}

\vspace{-2em}

\begin{figure}[H]
  \centering
  \includegraphics[width=\linewidth, height=0.9\textheight, keepaspectratio]{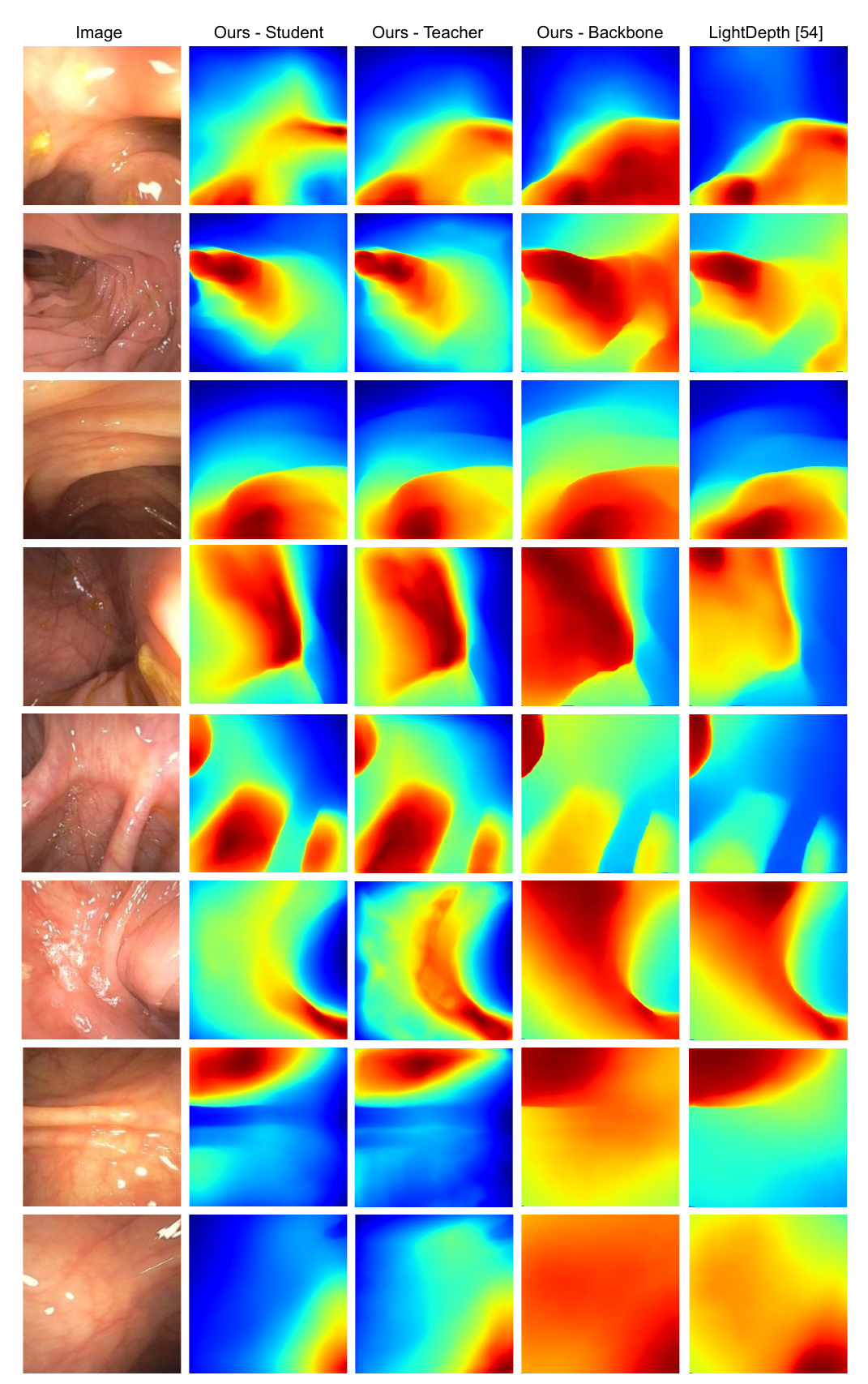}
  \caption{Qualitative evaluation on clinical data. Red = further distance from the camera and blue is closer.}
  \label{fig:clinical_results_full}
\end{figure}

\section{Limited Code Release}
\label{sec:supp_code}

Our limited code release can be accessed via a link to a GitHub repo on our project website: \url{https://ppsnet.github.io/}. The GitHub repo contains a \emph{PPSNet.py} file that serves as the full model file for our proposed approach. \emph{calculate\_PPL.py} from~\cite{lichy2022fast} is also included for reference alongside our proposed loss functions in \emph{PPS\_losses.py}. We also provide limited evaluation code for the C3VD~\cite{Bobrow_2023} dataset and pre-trained models. We will release our full codes, including the training code and various baselines, in the near future.
\section{LightDepth Implementation}
\label{sec:supp_lightdepth}
\vspace{-0.5em}
For the purpose of comparison to state-of-the-art monocular depth estimation specific to endoscopy, we attempted to implement LightDepth\cite{rodríguezpuigvert2023lightdepth} on our own, as the authors did not release their code. The authors describe an architecture with two ViT branches: one initialized with the weights from DPT-Hybrid, used for depth prediction; and one trained from scratch, used for prediction of albedo. Specifically, this albedo predictor computes hue and saturation values for each pixel, and then converts to RGB color space from HSV assuming V=100\%. We were unable to successfully implement this two-headed depth and albedo prediction. Instead, we compute the albedo $\rho$ by analytically solving the per-pixel rendering equation used (eq. 3 in \cite{rodríguezpuigvert2023lightdepth}):
\begin{align*}
    \mathcal{I}(d_i, \rho_i, g) &= \left(\frac{\sigma_0}{||d_i\mathbf{r}_i-\mathbf{x}_l||^2}R(\psi_i)\cos\theta_i\rho_i g\right)^{1/\gamma}
\end{align*}
with $\psi_i$, $R(\psi_i)$, and $\theta_i$ depending only on constants and the predicted depth map (whose value at pixel $i$ is $d_i$).

Assuming a colocated light and camera and setting $\sigma_0=g=1$, we can rearrange to get:
\begin{align*}
    \rho_i &= \frac{d_i^2 \mathcal{I}_i^\gamma}{R(\psi_i)\cos\theta_i}
\end{align*}

where $\mathcal{I}_i$ is the R, G, or B value of the $i$th pixel, and we apply this separately for each channel. This calculated $\rho_i$ replaced the predicted albedo $\rho_i$ in all loss calculations. We additionally add a regularization loss
\begin{align*}
    \mathcal{L}_{albvar} &= \frac{1}{3}(\mathrm{Var}(\rho^R) + \mathrm{Var}(\rho^G) + \mathrm{Var}(\rho^B))
\end{align*}
where $\mathrm{Var}(\rho^R)$ is the variance of the R channel calculated albedo value over the entire batch, and likewise for G and B. This provided a slight improvement in results, the idea being that in endoscopy the surface albedo should be fairly uniform.

Starting from the pretrained DPT-Hybrid weights from \cite{eftekhar2021omnidata}, we fine-tuned our LightDepth implementation for a single epoch on the train split of C3VD used in \cite{rodríguezpuigvert2023lightdepth} at a learning rate of 4e-6, as any further training was found to overfit and reduce accuracy.
\section{Qualitative Results on Bronchoscopy}
\label{sec:supp_bronchoscopy}

\begin{figure}[H]
  \centering
  \includegraphics[width=0.7\linewidth]{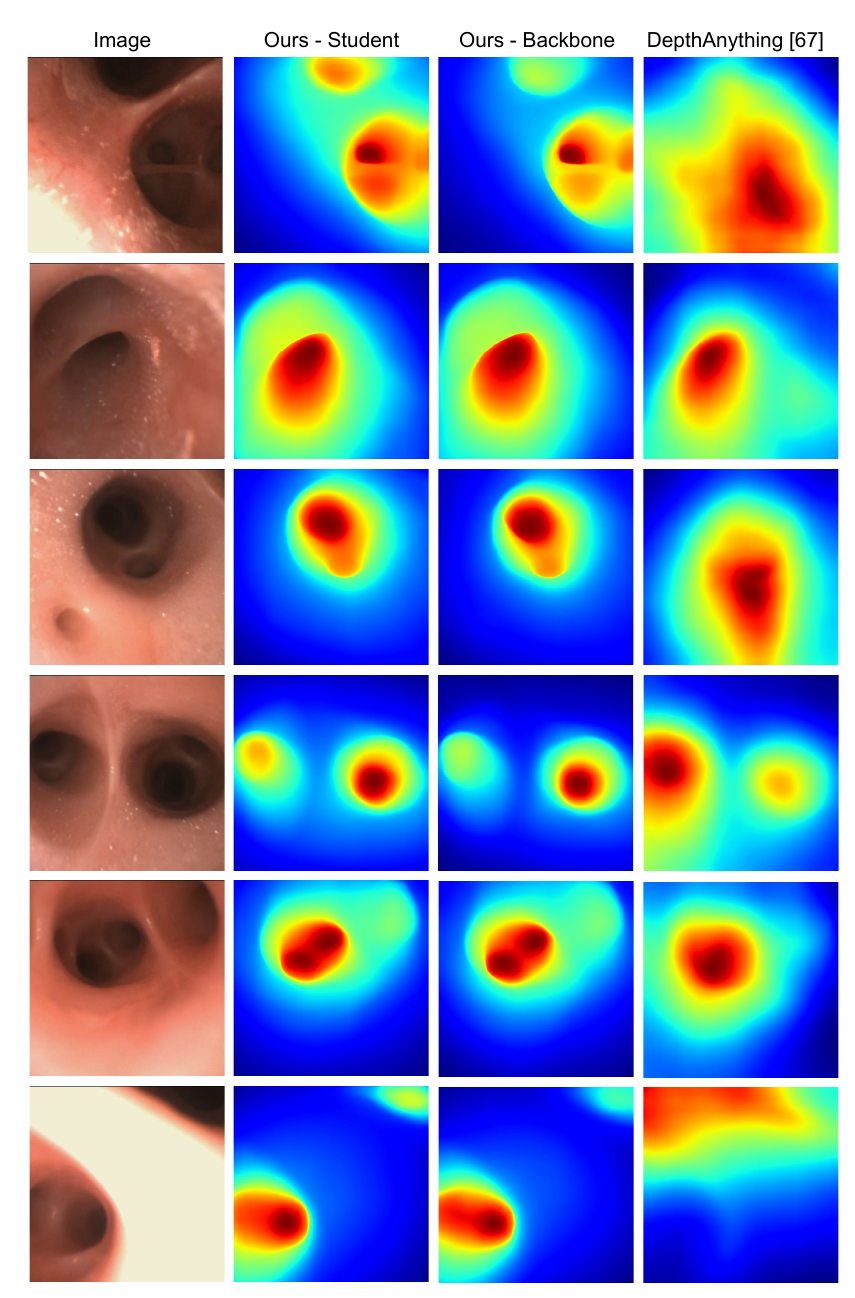}
  \caption{Qualitative evaluation on bronchoscopy data. Red = further distance from the camera and blue is closer. Note that 'Ours - Student' and 'Ours - Backbone' provide significantly higher quality depth estimations than Depth Anything~\cite{yang2024depth}.}
  \label{supp_fig:bronchoscopy_qualitative_results}
\end{figure}

In Fig.~\ref{supp_fig:bronchoscopy_qualitative_results}, we present qualitative depth map results on bronchoscopy data. Despite not on bronchoscopy data, we note that 'Ours - Student' and 'Ours - Backbone' both provide superior results to the baseline provided by Depth Anything~\cite{yang2024depth}, with 'Ours - Student' being slightly better than 'Ours - Backbone' in certain areas with relatively farther depths. We plan to explore bronchoscopy data as an additional bio-application in subsequent work. We will release a small dataset of bronchsocopy frames for qualitative evaluation upon this work's acceptance.

\section{Clinical Dataset}
\label{sec:supp_clinical_dataset}

The clinical dataset used in our work consists of 80 sequences with oblique views (7,293 frames), 14 sequences with en-face views (832 frames), and 20 sequences with down-the-barrel (axial) views (10,216 frames). The down-the-barrel (axial) views are from the Colon10K~\cite{ma2021colon10k} dataset. The oblique views and en-face views are from a dataset which has been released as a part of a separate work~\cite{wang2024structurepreservingimagetranslationdepth} and can be found here: \url{https://endoscopography.web.unc.edu/place-recognition-in-colonoscopy/}.

%
%
\bibliographystyle{splncs04}
\bibliography{main}
\end{document}